# Predicting Alzheimer's Disease Diagnosis Risk over Time with Survival Machine Learning on the ADNI Cohort

Henry Musto[1], Daniel Stamate[1,2], Ida Pu[1], Daniel Stahl[3]

[1] Data Science & Soft Computing Lab, Computing Department,
Goldsmiths College, University of London, United Kingdom
[2] Division of Population Health, Health Services Research & Primary Care,
School of Health Sciences, University of Manchester, United Kingdom
[3] Institute of Psychiatry Psychology and Neuroscience, Biostatistics and Health Infomatics Department, King's College London, London, UK

**Abstract.** The rise of Alzheimer's Disease worldwide has prompted a search for efficient tools which can be used to predict deterioration in cognitive decline leading to dementia. In this paper, we explore the potential of survival machine learning as such a tool for building models capable of predicting not only deterioration but also the likely time to deterioration. We demonstrate good predictive ability (0.86 C-Index), lending support to its use in clinical investigation and prediction of Alzheimer's Disease risk.

## 1   Introduction

One of the most pressing challenges for governments and healthcare systems is the rising number of people with dementia. More than 55 million people live with dementia worldwide, and there are nearly 10 million new cases yearly, with 60-70% of all dementias being of Alzheimer's Disease type (AD) [1]. Recently, attention has turned to Machine Learning (ML) as a tool for improving the predictive ability of clinical models concerning AD and addressing clinical challenges more widely. However, of the hundreds of clinical ML models that appear in scientific publications each year, few have thus far been successfully embedded into existing clinical practice [2]. One of the reasons for this is that most models only provide predictions for disease cases without quantifying the probability of disease occurrence. This limitation restricts clinicians' ability to accurately measure and communicate the probability of disease development over time with the patient. [3]. Also, in the context of predicting the progression of AD in particular, many studies that use ML methods employ a classification approach, whereby the outcome to be predicted is either a binomial or multinomial outcome within a specific timeframe [4] [5]. The datasets are often derived from longitudinal studies, whereby clinical marker data is collected from participants over months and years [6]. Thus, such data has a temporal element inherent to the methodology employed in the collection process. However, standard classification ML cannot consider the predictive power of time in conjunction with other predictors. Furthermore, classification models cannot handle drop-outs which are common in longitudinal studies.

With this in mind, a newly emerging field of exploration seeks to build on traditional time-dependent statistical models, such as survival analysis, to develop machine learning models which can predict the time-dependent risk of developing AD and go beyond simple classification. Survival analysis is a statistical method that aims to predict the risk of an event's occurrence, such as death or the emergence of a disease, as a function of time. A key aspect of survival analysis is the presence of censored data, indicating that the event of interest has not occurred while the subject was part of the study. The presence of censored data requires the use of specialised techniques. Traditionally, the Cox proportional hazards model [7] has been the most widely used technique for analysing data containing also censored records. However, the Cox model typically works well for small data sets and does not scale well to high dimensions [8]. ML techniques that inherently handle high-dimensional data have been adapted to handle censored data, allowing ML to offer a more flexible alternative for analysing high-dimensional, censored, heterogeneous data [8]. Furthermore, the ability to predict not only a binary or multinomial outcome but also the risk of such outcomes occurring at different timepoints provides clinicians and researchers with more information for the benefit of research and patients.

This work has several aims. First, it aims to build upon existing work demonstrating the utility of survival-based ML techniques in predicting the risk of deterioration at different time points in AD using the Alzheimer's Disease Neuroimaging Initiative (ADNI) database. Secondly, it aims to explore the predictive power of these techniques once the more physically intrusive biomarkers available in the dataset are removed. These predictors, such as ABETA, TAU and PTAU, which are established biomarkers for dementia, are collected via painful lumbar puncture procedures to sample cerebrospinal fluid (CSF). Recently efforts have been made to investigate alternative biomarkers such as blood metabolites which, in some studies, proved to have comparable predictive power to the established CSF-biomarkers [9].

The rest of the paper will be ordered as follows. First, it will review existing literature on survival-based ML as applied to clinical questions in general and AD prediction in particular. Next, the problem of interest will be defined. Then the proposed methodology will be outlined. Before the results are presented, the study design of the dataset will be described, including predictors and diagnostic criteria. A discussion of the implications of these results will then follow.

## 2  Related Work

Spooner et al. [8] systematically compared the performance and stability of ML algorithms and feature selection methods suitable for high-dimensional, heterogeneous, censored clinical data, in the context of cognitive ageing and AD, by predicting the risk of AD over time [8]. The authors assessed ten survival-based machine-learning techniques alongside the standard Cox proportional hazard model. The Sydney Memory and Aging Study (MAS) dataset and Alzheimer's Disease Neuroimaging Initiative (ADNI) dataset were utilised. All algorithms evaluated performed well on both data sets and outperformed the standard Cox proportional hazards model.

Another paper that explores the clinical utility of survival modelling within the domain of AD research comes from [10], which looked at the interaction between socioeconomic features and polygenic hazard scores on the timing of Alzheimer's diagnosis using Cox proportional hazard survival analysis. Only the standard Cox PH technique was used. The authors could demonstrate the clinical utility of using socioeconomic markers and the presence of the APOE4 gene expression to predict the time to AD diagnosis. Although a small study focusing on only one model, this work demonstrated the utility of survival-based models in AD prediction. However, more work was needed to build upon these results using ML methods. This was achieved in [11] using ML survival-based methods to predict the risk of developing AD in the English Longitudinal Study of Aging (ELSA) dataset. This work again found that Survival ML outperformed Cox methods.

On the other hand, [12] found the standard Cox regression and two ML models (Survival Random Forest and Extreme Gradient Boosting) had comparable predictive accuracy across three different performance metrics, when applied to the Prospective Registry For Persons with Memory Symptoms (PROMPT) dataset [13]. The authors concluded that survival ML did not perform better than standard survival methods.

In comparison, [14] found that multi-modal survival-based deep learning methods produced good results when applied to the ADNI dataset, comparable to [8]. In this context, our present work serves as an example of including neural network models, as these methods have hitherto seldom been explored in a survival context.

Despite the scarcity of survival modelling papers in relation to AD prediction, recent examples have shown promise in attempting to outperform the classic Cox proportional hazard model, using survival ML and survival neural networks/ deep learning on clinical datasets. This supports the continued exploration of survival ML as a predictive tool for clinical risk problems [11].

## 3   Problem Definition

This study uses survival-based ML methods to predict the risk of deterioration, defined as receiving a worse diagnosis at their final visit to the data collection centre before leaving the study, compared to baseline diagnosis. Furthermore, the study aims to build models to predict the risk of receiving a worse diagnosis within the data collection period using survival-based ML. These models will then be tested for stability, and two estimations of the general test error will be calculated based on C-Index and Calibration scores [15].

A secondary aim is to explore the predictive power of these models when predictors derived from invasive CSF collections are removed from the dataset.

## 4   Methodology
### 4.1   Data Description

**Alzheimer's Disease Neuroimaging Initiative.**
The data used in this paper was derived from the Alzheimer's Disease Neuroimaging Initiative (ADNI) database [6]. This longitudinal case-control study was initiated in 2004 by the National Institute of Aging (NIA), The National Institute of Biomedical Imaging and Bioengineering (NIBIB), The Food and Drug Administration (FDA), as well as elements of the private and non-profit sectors. The initial protocol, ADNI1, was conducted over six years, recruiting 400 subjects diagnosed with Mild Cognitive Impairment (MCI), 200 subjects with Alzheimer's (AD), and 200 healthy controls (CN). The initial goal of the ADNI study was to test whether repeated collections of neuroimaging, biomarker, genetic, and clinical and neuropsychological data could be combined to contribute in an impactful way to research dementia [6].

Data for the present paper was downloaded on the 1st of October 2022 through the ADNIMERGE package in R. This package combines predictors from the different ADNI protocols. The final combined dataset contains 115 variables and 15,157 observations, which included multiple observations per participant. These observations represent data collection events where participants made up to 23 visits to study sites. The data used for this work is a subset of the full dataset, containing only information from the original ADNI2 study. After some initial cleaning, the resulting data contained 607 observations and 52 variables consisting of 50 input attributes, 1 time attribute (defined as the time in months until the participant visited the data collection centre for the last time), and 1 outcome attribute. The outcome attribute consisted of three diagnostic classes received at their final visit to the data collection centre: those who received a diagnosis of Cognitively Normal (CN), those who received a diagnosis of Mild Cognitive Impairment (MCI), and those who received a diagnosis of Alzheimer's Disease (AD) [4].

### 4.2 Predictors

- Baselines Demographics: age, gender, ethnicity, race, marital status, and education level were included in the original dataset.

- Neuropsychological test results, including those from the Functional Activities Questionnaire (FAQ), the Mini-Mental State Exam (MMSE), and Rey's Auditory Verbal Learning Test (RAVLT), were included in the data. This numeric data is well-validated as a tool for identifying cognitive impairment in general and AD-related cognitive impairment in particular. Full details of the tests included can be found at [16].

- Positron Emission Tomography (PET) measurements (FDG, PIB, AV45) are indirect measures of brain function using the Positron Emission Tomography neuroimaging modality.

- Magnetic Resonance Imaging (MRI) measurements (Hippocampus, intracranial volume (ICV), MidTemp, Fusiform, Ventricles, Entorhinal and WholeBrain) are structural measurements of a participant's brain derived from the Magnetic Resonance Imaging neuroimaging modality.

- APOE4 is an integer measurement representing the appearance of the epsilon 4 allele of the APOE gene. This allele has been implicated as a risk factor for AD [17]

- ABETA, TAU, and PTAU are cerebrospinal fluid (CSF) biomarker measurements. These biomarkers are collected via lumbar puncture. These predictors were removed from the model-building process for the second set of models.

- Last Visit is defined for this paper as the number of months from baseline data collection to the subject's last visit at a data collection centre. This variable was added to explicitly define a time predictor for survival-based ML modelling.

### 4.3 Data Preprocessing

Boolean variables were created, indicating the location of missing data for each predictor. Variables with missingness at 90% or greater of the total rows for that predictor were removed. All nominal predictors were dummy-coded.

The data was split into two groups to predict deterioration using survival-based ML. The first group contained only those diagnosed as cognitively normal (CN) on their first visit to the data collection centre. The second group contained only those diagnosed with Mild Cognitive Impairment (MCI) on their first visit to the data collection centre. Deterioration was defined as receiving a worse diagnosis on their final visit to the data collection centre. The resultant two datasets had 285 and 322 observations respectively and 98 variables with CSF-derived biomarkers included/92 without (See Tables 1, 2, 3).

**Table 1.** Those who received a cognitively normal (CN) diagnosis at baseline were the only group included. The models predicted the diagnoses these participants received at the final visit, defined here.

| Outcome | Definition |
|---------|------------|
| CN | Those diagnosed with CN at baseline who received the same diagnosis at their last visit. |
| MCI/AD | Those having received a diagnosis of CN at baseline *either* received a diagnosis of AD or MCI at their last visit. |

**Table 2.** Those diagnosed with Mild Cognitive Impairment (MCI) at baseline were the only group included. The models predicted the diagnoses these participants received at the final visit, defined here.

| Out-come | Definition |
|---|---|
| CN/MCI | Those who had received a diagnosis of MCI at baseline either received the same diagnosis at their last visit or a more favourable diagnosis of CN. |
| AD | Those diagnosed with MCI at baseline received a diagnosis of AD at their last visit. |

**Table 3.** The final dimensions of the two datasets after preprocessing.

| Dataset | Variables | Observations |
|---|---|---|
| CN at baseline | 98/92 (with/without CSF predictors) | 285 |
| MCI at baseline | 98/92 (with/without CSF predictors) | 322 |

### 4.4 Model Development

Model development, evaluation, and validation were carried out according to methodological guidelines outlined by [18]; results were reported according to the Transparent Reporting of a multivariable prediction model for Individual Prognosis or Diagnosis (TRIPOD) guidelines [19]. This paper explored three algorithms:

Cox Proportional Hazard Model (Cox PH) - The Cox model is expressed by the hazard function, which is the risk of an event occurring at time as follows:

$$h(t) = h_0(t) * \exp(\beta_1 X_1 + \beta_2 X_2 + \beta_p X_p) \qquad (1)$$

where $t$ represents the survival time, $h(t)$ is the hazard function, $X_1, X_2, ... X_p$ are the values of the $p$ covariates, $\beta_1, \beta_2 ... \beta_p$ are the coefficients that measure the effect of the covariates on the survival time and $h_0(t)$ is the baseline hazard function, which is unspecified. The regression coefficients are estimated by maximising the partial likelihood [8], and hence the model does not require tuning.

Survival Random Forest (SRF) - Random Forests seek to grow many trees using bootstrapped aggregation and splitting on a random subsection of predictors for each split point. The split points are chosen based on some criteria (such as entropy or purity of the node), which seeks to allocate classifications of one type within each terminal node. In a Survival Random Forest, the feature and split point chosen is the one that maximises the survival difference (in terms of the hazard function) between subsequent nodes [8] [20]. In the tuning grid for this model, the values of mtry varied between 1 and 20, with a step of 1, while the values for minimum node size in the grid were 10, 20, 30, 40, and 50. SRF comprised 1000 trees. The number of trees promotes model convergence (large is better) and generally is not tuned.

Survival Deep Hit Neural Networks (SNN) - Deep Hit is a multi-task neural network comprising a shared sub-network and K cause-specific sub-networks. The architecture differs from a conventional multi-task neural network in two ways. First, it utilises a single softmax layer as the output layer of Deep Hit to ensure that the network learns the joint distribution of K possible outcomes, not the marginal distributions of each outcome. Second, it maintains a residual connection from the input covariates into the input of each cause-specific sub-network. The full technical description of this model can be found in [21]. In the tuning grid for this model, the number of nodes was between 2 and 300, the epochs were between 10 and 400, and the batch sizes was 32. The learning rates were 0.001, and 0.01, the activation functions were 'relu', 'elu' and 'leakyrelu', and the optimisers were 'adam' and 'adamw'. 10% of the training dataset was held aside for validation in the early stopping procedure, with patience at either 10, or 150 epochs.

### 4.5 Nested Cross-Validation and Monte Carlo Simulation

A Nested Cross-Validation procedure was implemented to tune and evaluate the models so precise estimates of the model's performance of unseen cases (internal validation) could be gathered [4]. Nested Cross-Validation consisted of an outer 5-fold CV (model assessment) and an inner 5-fold CV (model tuning). We conducted a Monte Carlo procedure of 100 repetitions of the nested CV using different random splits per model to assess the models' stability. Performance statistics were recorded for each model produced by each iteration. Each performance statistic's mean and standard deviation across all iterations were recorded when the MC was complete. To ensure the representativeness of training and test samples in both procedures, the data splitting was stratified based on the AD cases variable.

### 4.6 Performance Metrics

To assess model performance, two statistics were recorded. Discrimination was assessed using the Concordance index or C-index [18]. This metric, also called Harrel's C-index, provides a global assessment of the model and can be considered a more general form of the AUCROC measure typically used in binary classification tasks. The C-index computes the percentage of comparable pairs within the dataset whose risk score was correctly identified by the model. Comparable pairs are defined as a selection of two observations, which can be compared in terms of survival time predicted by the model. If both are censored, then they are not included in the computation for this metric. A pair is considered concordant if the observation who experiences the earlier event is identified as having greater risk and discordant otherwise. Thus the total concordance score for a model is the ratio of concordant pairs within the dataset divided by the total number of observations [15].

Secondly, calibration was assessed using Van Houwelingen's Alpha Survival Measure of non-proportional hazards models [15]. This metric is defined as:

$$\alpha = \sum \delta / \sum H_i(t_i) \quad (2)$$

Where $\delta$ is the true censoring indicator observed from the test data, $H_i$ is the cumulative hazard predicted by the model, and $t_i$ is the observed survival time. The model is well calibrated if the estimated α is equal or close to 1. Calibration is a formal comparison between the probability distribution and resultant survival instances observed in the test data and the probability distribution and resultant survival predictions generated by the model. A full exploration of this metric can be found in [22].

### 4.7    Software and Hardware

The data analysis was conducted using the R language [23]. Initial data cleaning was performed using base R functions and the Tidyverse R package [24]. The creation of dummy variables was performed using the Caret R package [25]. The nested cross-validation procedure, including training, tuning and evaluation, was performed on the Cox PH, SRF, and SNN models using the mlr3 R package [26]. The hardware consisted of 3 servers running Linux, with Xeon processors and 64GB of RAM.

## 5    Results

The nested cross-validation C-index and Calibration performance for each model type is detailed below. Figures for the two groups' C-indexes, with CSF-derived biomarkers included in the models, can be found in Fig. 2.

**Table 4.** CN group with CSF-derived biomarkers included / removed.

| Model | C-index CSF included / removed | Calibration CSF included / removed |
|---|---|---|
| Cox PH | 0.71 / 0.59 | 0.01 / 0.01 |
| SRF | 0.84 / 0.86 | 0.80 / 1.02 |
| SNN | 0.80 / 0.70 | 0.64 / 0.60 |

The best-performing model for the CN group with CSF-derived biomarkers included was SRF, followed by SNN, followed by Cox PH model. Thus, the SRF and SNN outperformed the conventional statistical model Cox in the CN group with CSF-derived biomarkers included in the Calibration and the C-index metric.

Once the CSF-derived biomarkers were removed, for the CN group, both the Cox PH and the SNN reported worse predictive power. However, as the C-Index and Calibration estimated, the SRF retained its predictive ability, even significantly improving its calibration score.

**Table 5.** MCI group with CSF-derived biomarkers included / removed.

| Model | C-index CSF included / removed | Calibration CSF included / removed |
|---|---|---|
| Cox PH | 0.78 / 0.78 | 0.29 / 0.25 |
| SRF | 0.84 / 0.84 | 0.98 / 0.99 |
| SNN | 0.83 / 0.77 | 1.16 / 0.91 |

When considering the C-index, the best-performing model for the MCI group, with CSF-derived biomarkers included, was SRF, followed by Cox PH model, followed by SNN. Calibration was again almost perfect for SRF followed by SNN and CoxPH.

Once the CSF-derived biomarkers were removed, for the MCI group, only the SNN reported worse predictive power, as measured by the C-Index. When considering calibration, however, the SNN and Cox PH models deteriorated when the CSF-derived biomarkers were removed, while SRF remained close to 1.

The datasets with the CSF-derived biomarkers removed were then taken forward for all models to undergo a Monte Carlo simulation with 100 iterations of the nested cross-validation procedure.

**Table 6.** Cox PH Monte Carlo at 100 iterations.

| Group (Model) | Mean C-index (sd) | Mean Calibration (sd) |
|---|---|---|
| MCI (Cox PH) | 0.78(0.02) | 0.33(0.08) |
| CN (Cox PH) | 0.59(0.06) | 0.03(0.02) |

**Table 7.** SNN Monte Carlo.

| Group | Mean C-index (sd) | Mean Calibration (sd) |
|---|---|---|
| MCI (SNN) | 0.77(0.02) | 0.91(0.1) |
| CN (SNN) | 0.7(0.06) | 0.6(0.03) |

**Table 8.** SRF Monte Carlo.

| Group | Mean C-index (sd) | Mean Calibration (sd) |
|---|---|---|
| MCI (SRF) | 0.84(0.008) | 0.99(0.02) |
| CN (SRF) | 0.83(0.01) | 1.02(0.02) |

The SRF model results on both the C-Index and Calibration proved the most stable upon repeated testing, with standard deviations at less than 0.03. The SNN model was less stable and reported less predictive power, as measured by both the C-Index and Calibration.

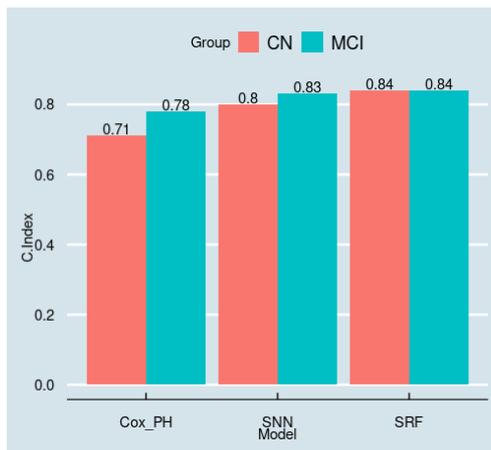

**Fig. 1.** C-indexes for models applied to the two groups with CSF-derived biomarkers included in the models.

## 6   Discussion

This study aimed to further explore the potential of survival-based ML as a tool for predicting time to AD diagnosis. This paper demonstrates the clear utility of such methods when predicting on the ADNI2 dataset. This provides further evidence for the continued exploration of the utility of survival ML in this context.

Several results reported here are worthy of note. Firstly, we demonstrated good predictive power for SRF with very good discrimination and excellent calibration, which was superior to both the standard Cox PH model and the SNN model. Good discrimination and calibration are essential in survival ML models to obtain accurate risk estimations at specific time periods of interest, which is not possible with traditional classification ML models. This allows for informed decision-making, personalised interventions, and timely allocation of resources for the prevention, early detection, or management of dementia. Our results support the work of [11] but disagrees with [20], which found that the standard Cox model was superior to tree-based ensemble methods. This is possibly due also to the way in which the Survival trees were constructed, with [18] using probabilities derived from a Cox model to construct a Random Forest. In comparison, the SRF presented here sought to create trees whose splits aimed to maximise the difference in survival between the resultant nodes. With the present study indicating strong results using this approach, it may be that the latter technique produces better models. However, we should note that these results were obtained on datasets other than the one used in this study, ADNI.

With the removal of the CSF-derived biomarkers, performance deterioration was seen for SNN but not SRF or the Cox PH. The choice to investigate an SNN was derived, in part, from the work of [14], whose best model achieved a C-index of 0.83 on the ADNI dataset. In comparison, the best model found by the present study, using SNN, achieved a C-index of 0.77. However, we should note that [13] did not provide a comparison between the Survival Neural Network models used and either a standard Cox PH model or any other survival ML algorithm. Another point of consideration is that the authors used a slightly different Neural Network algorithm to the one described here. Thus, an important next step would be directly comparing the DeepSurv model and the Deep Hit model described here.

SNN had worse stability than the SRF and Cox PH models, as measured by the standard deviations of the C-index and Calibration scores for these models. This would suggest that this algorithm produces unstable models with unreliable predictions. Neural Networks usually perform best in complex problems that require discovering hidden patterns in the data between a large number of interdependent variables. Furthermore, Neural Networks usually perform better on image and audio classification rather than tabular data, such as the dataset used in this study [27]. Therefore, it may be the case that a simpler model such as Random Forest might be better suited for the kind of limited datasets presented here. It may also be the case that the SNN model overfit the comparatively small dataset presented here.

Finally, the results in this work suggest that CSF-derived biomarkers did not have a clear contribution in this setting, for building models capable of accurately predicting the time to AD diagnosis on our considered ADNI sample. Although both the Cox PH and SNN models variously suffered from the removal of these predictors, the RSF model did not. This is important, as collecting biomarkers from CSF is an invasive and painful process for participants, which involves a lumbar puncture. Recent analyses conducted on EMIF-AD data [9] established that predictors such as metabolites in blood showed similar predictive power to the well-established but more invasive CSF biomarkers.

Despite the results obtained by this work, there are a number of limitations to the present paper that need to be considered. Firstly, the ADNI2 data is comparatively small, and future work is required to validate the models created here using external data. A related point is the lack of diversity within this data, which heavily skews towards white North-American participants. To validate the models created here, they must be tested on non-white, non-western participants such that evidence of model performance be gathered for a wider group of people.

A further limitation is that the choice of hyper-parameters for the grid search procedure for each model is finite. We were unable to conduct an exhaustive search over a larger set of combinations of hyperparameter values due to time constraints and computational cost. Therefore it is entirely possible that better results for these models can be found using hyperparameters not explored here.

## 7      Conclusion

This paper proposed a survival ML approach to predict the time to Alzheimer's Disease diagnosis accurately. It was compared with one of the most used statistical models for survival analysis, namely Cox PH. In our framework proposed by using the ADNI cohort, the Machine Learning based approach proved to be more accurate than the statistical approach, which was the case also in a recent study conducted on different clinical data [11].

**Acknowledgements**
Daniel Stahl was part funded by the NIHR Maudsley Biomedical Research Centre at South London and Maudsley NHS Foundation Trust and King's College London. This study represents independent research and views expressed are those of the author(s) and not necessarily those of the NIHR or the Department of Health and Social Care.